\documentclass[11pt]{article}
\usepackage{graphicx,amsmath,amsfonts,amssymb,bm,hyperref,url,breakurl,epsfig,epsf,color,fullpage,MnSymbol,mathbbol, fmtcount,algorithmic,algorithm, semtrans
} 

\usepackage{titlesec}

\usepackage{tikz}
\usepackage{pgfplots}

\usetikzlibrary{pgfplots.groupplots}

\titleformat{\paragraph}
{\normalfont\normalsize\bfseries}{\theparagraph}{1em}{}
\titlespacing*{\paragraph}
{0pt}{3.25ex plus 1ex minus .2ex}{1.5ex plus .2ex}

\usepackage{caption,subcaption}

\usepackage[bottom,hang,flushmargin]{footmisc} 

\usepackage{hyperref}
\definecolor{darkred}{RGB}{150,0,0}
\definecolor{darkgreen}{RGB}{0,150,0}
\definecolor{darkblue}{RGB}{0,0,200}
\hypersetup{colorlinks=true, linkcolor=darkred, citecolor=darkgreen, urlcolor=black}

\newtheorem{theorem}{Theorem}[section]
\newtheorem{lemma}[theorem]{Lemma}
\newtheorem{corollary}[theorem]{Corollary}

\newtheorem{definition}[theorem]{Definition}

\newcommand{\tn}[1]{\left\|#1\right\|_{\ell_2}}
\newcommand{\Cc}{\mathcal{C}}

\newcommand{\Bc}{\mathcal{B}}

\newcommand{\des}{{\vct{w}}}

\newcommand{\h}{\vct{h}}

\newcommand{\twonorm}[1]{\left\|#1\right\|_{\ell_2}}

\newcommand{\infnorm}[1]{\left\|#1\right\|_{\ell_\infty}}

\newcommand{\abs}[1]{\left|#1\right|}

\newcommand{\R}{\mathbb{R}}

\newcommand{\sgn}[1]{\textrm{sgn}(#1)}

\newcommand{\E}{\operatorname{\mathbb{E}}}

\newcommand{\vct}[1]{\bm{#1}}
\newcommand{\mtx}[1]{\bm{#1}}

\definecolor{ejc}{RGB}{0,0,255}

\numberwithin{equation}{section} 

\def \endprf{\hfill {\vrule height6pt width6pt depth0pt}\medskip}
\newenvironment{proof}{\noindent {\bf Proof} }{\endprf\par}

\title{Learning ReLUs via Gradient Descent}
\author{Mahdi Soltanolkotabi\\Ming Hsieh Department of Electrical Engineering\\University of Southern California, Los Angeles, CA, 90089}
\date{May 2017}
\begin{document}
\maketitle
\begin{abstract} 
In this paper we study the problem of learning Rectified Linear Units (ReLUs) which are functions of the form $\vct{x}\mapsto \max(0,\langle \vct{w},\vct{x}\rangle)$ with $\vct{w}\in\R^d$ denoting the weight vector.  We study this problem in the high-dimensional regime where the number of observations are fewer than the dimension of the weight vector. We assume that the weight vector belongs to some closed set (convex or nonconvex) which captures known side-information about its structure. We focus on the realizable model where the inputs are chosen i.i.d.~from a Gaussian distribution and the labels are generated according to a planted weight vector. We show that projected gradient descent, when initialization at $\vct{0}$, converges at a linear rate to the planted model with a number of samples that is optimal up to numerical constants. Our results on the dynamics of convergence of these very shallow neural nets may provide some insights towards understanding the dynamics of deeper architectures.
\end{abstract}

\section{Introduction}

Nonlinear data-fitting problems are fundamental to many supervised learning tasks in signal processing and machine learning. Given training data consisting of $n$ pairs of input features $\vct{x}_i\in\R^d$ and desired outputs $\vct{y}_i\in\R$ we wish to infer a function that best explains the training data. In this paper we focus on fitting Rectified Linear Units (ReLUs) to the data which are functions $\phi_{\vct{w}}: \R^d\rightarrow\R$ of the form
\begin{align*}
\phi_{\vct{w}}(\vct{x})=\max\left(0,\langle \vct{w},\vct{x}\rangle\right).
\end{align*}
A natural approach to fitting ReLUs to data is via minimizing the least-squares misfit aggregated over the data. This optimization problem takes the form
\begin{align}
\label{opt}
\underset{\vct{w}\in\R^d}{\min}\quad \mathcal{L}(\vct{w}):=\frac{1}{n}\sum_{i=1}^n \left(\max\left(0,\langle \vct{w},\vct{x}_i\rangle\right)-y_i\right)^2\quad\text{subject to}\quad \mathcal{R}(\vct{w})\le R,
\end{align}
with $\mathcal{R}:\R^d\rightarrow\R$ denoting a regularization function that encodes prior information on the weight vector. 

Fitting nonlinear models such as ReLUs have a rich history in statistics and learning theory \cite{kalai2009isotron} with interesting new developments emerging \cite{goel2016reliably} (we shall discuss all these results in greater detail in Section \ref{mysec}). Most recently, nonlinear data fitting problems in the form of neural networks (a.k.a. deep learning) have emerged as powerful tools for automatically extracting interpretable and actionable information from raw forms of data, leading to striking breakthroughs in a multitude of applications \cite{krizhevsky2012imagenet, mohamed2012acoustic, collobert2008unified}. In these and many other empirical domains it is common to use local search heuristics such as gradient or stochastic gradient descent for nonlinear data fitting. These local search heuristics are surprisingly effective on real or randomly generated data. However, despite their empirical success the reasons for their effectiveness remains mysterious. 


Focusing on fitting ReLUs, a-priori it is completely unclear why local search heuristics such as gradient descent should converge for problems of the form \eqref{opt}, as not only the regularization function maybe nonconvex but also the loss function!  Efficient fitting of ReLUs in this high-dimensional setting poses new challenges: When are the iterates able to escape local optima and saddle points and converge to global optima? How many samples do we need? How does the number of samples depend on the a-priori prior knowledge available about the weights? What regularizer is best suited to utilizing a particular form of prior knowledge?  How many passes (or iterations) of the algorithm is required to get to an accurate solution?
At the heart of answering these questions is the ability to predict convergence behavior/rate of (non)convex constrained optimization algorithms. In this paper we build up on a new framework developed by the author in \cite{soltanolkotabi2017structured} for analyzing nonconvex optimization problems to address such challenges.

\section{Precise measures for statistical resources}
We wish to characterize the rates of convergence for the projected gradient updates \eqref{iters} as a function of the number of samples, the available prior knowledge and the choice of the regularizer. To make these connections precise and quantitative we need a few definitions. Naturally the required number of samples for reliable data fitting depends on how well the regularization function $\mathcal{R}$ can capture the properties of the weight vector $\vct{w}$. For example, if we know that the weight vector is approximately sparse, naturally using an $\ell_1$ norm for the regularizer is superior to using an $\ell_2$ regularizer. To quantify this capability we first need a couple of standard definitions which we adapt from \cite{oymak2015sharp, oymak2016fast, soltanolkotabi2017structured}.
\begin{definition}[Descent set and cone] \label{decsetcone} The \emph{set of descent} of  a function $\mathcal{R}$ at a point $\vct{w}^*$ is defined as
\begin{align*}
{\cal D}_{\mathcal{R}}(\vct{w}^*)=\Big\{\vct{h}:\text{ }\mathcal{R}(\vct{w}^*+\vct{h})\le \mathcal{R}(\vct{w}^*)\Big\}.
\end{align*}
The \emph{cone of descent} is defined as a closed cone $\mathcal{C}_{\mathcal{R}}(\vct{w}^*)$ that contains the descent set, i.e.~$\mathcal{D}_{\mathcal{R}}(\vct{w}^*)\subset\mathcal{C}_{\mathcal{R}}(\vct{w}^*)$. The \emph{tangent cone} is the conic hull of the descent set. That is, the smallest closed cone $\mathcal{C}_{\mathcal{R}}(\vct{w}^*)$ obeying $\mathcal{D}_{\mathcal{R}}(\vct{w}^*)\subset\mathcal{C}_{\mathcal{R}}(\vct{w}^*)$.
\end{definition}
We note that the capability of the regularizer $\mathcal{R}$ in capturing the properties of the unknown weight vector $\vct{w}^*$ depends on the size of the descent cone $\mathcal{C}_{\mathcal{R}}(\vct{w}^*)$. The smaller this cone is the more suited the function $\mathcal{R}$ is at capturing the properties of $\vct{w}^*$. To quantify the size of this set we shall use the notion of mean width.
\begin{definition}[Gaussian width]\label{Gausswidth} The Gaussian width of a set $\mathcal{C}\in\R^d$ is defined as:
\begin{align*}
\omega(\mathcal{C}):=\mathbb{E}_{\vct{g}}[\underset{\vct{z}\in\mathcal{C}}{\sup}~\langle \vct{g},\vct{z}\rangle],
\end{align*}
where the expectation is taken over $\vct{g}\sim\mathcal{N}(\vct{0},\mtx{I}_p)$. Throughout we use $\mathcal{B}^d/\mathbb{S}^{d-1}$ to denote the the unit ball/sphere of $\R^d$.
\end{definition}
We now have all the definitions in place to quantify the capability of the function $\mathcal{R}$ in capturing the properties of the unknown parameter $\vct{w}^*$. This naturally leads us to the definition of the minimum required number of samples.
\begin{definition}[minimal number of samples]\label{PTcurve}
Let $\mathcal{C}_{\mathcal{R}}(\vct{w}^*)$ be a cone of descent of $\mathcal{R}$ at $\vct{w}^*$. We define the minimal sample function as
\begin{align*}
\mathcal{M}(\mathcal{R},\vct{w}^*)=\omega^2(\mathcal{C}_{\mathcal{R}}(\vct{w}^*)\cap\mathcal{B}^d).
\end{align*}
We shall often use the short hand $n_0=\mathcal{M}(\mathcal{R},\vct{w}^*)$ with the dependence on $\mathcal{R},\vct{w}^*$ implied.  
\end{definition}
We note that $n_0$ is exactly the minimum number of samples required for structured signal recovery from linear measurements when using convex regularizers \cite{Cha, McCoy}. Specifically, the optimization problem
\begin{align}
\label{lininv}
\sum_{i=1}^n \left(y_r-\langle \vct{x}_i,\vct{w}^*\rangle\right)^2\quad\text{subject to}\quad \mathcal{R}(\vct{w})\le\mathcal{R}(\vct{w}^*),
\end{align}
succeeds at recovering an unknown weight vector $\vct{w}^*$ with high probability from $n$ observations of the form $\vct{y}_i=\langle \vct{a}_i,\vct{w}^*\rangle$ if and only if $n\ge n_0$.\footnote{We would like to note that $n_0$ only approximately characterizes the minimum number of samples required. A more precise characterization is $\phi^{-1}(\omega^2(\mathcal{C}_{\mathcal{R}}(\vct{w}^*)\cap\mathcal{B}^d))\approx \omega^2(\mathcal{C}_{\mathcal{R}}(\vct{w}^*)\cap\mathcal{B}^d)$ where $\phi(t)=\sqrt{2}\frac{\Gamma\left(\frac{t+1}{2}\right)}{\Gamma\left(\frac{t}{2}\right)}\approx\sqrt{t}$. However, since our results have unspecified constants we avoid this more accurate characterization.} While this result is only known to be true for convex regularization functions we believe that $n_0$ also characterizes the minimal number of samples even for nonconvex regularizers in \eqref{lininv}. See \cite{oymak2015sharp} for some results in the nonconvex case as well as the role this quantity plays in the computational complexity of projected gradient schemes for linear inverse problems. Given that with nonlinear samples we have less information (we loose some information compared to linear observations) we can not hope to recover the weight vector from $n\le n_0$ when using \eqref{opt}. Therefore, we can use $n_0$ as a lower-bound on the minimum number of observations required for projected gradient descent iterations \eqref{iters} to succeed at finding the right model.


\section{Theoretical results for learning ReLUs}
A simple heuristic for optimizing \eqref{opt} is to use gradient descent. One challenging aspect of the above loss function is that it is not differentiable and it is not clear how to run projected gradient descent. However, this does not pose a fundamental challenge as the loss function is differentiable except for isolated points and we can use the notion of generalized gradients to define the gradient at a non-differentiable point as one of the limit points of the gradient in a local neighborhood of the non-differentiable point. For the loss in \eqref{opt} the generalized gradient takes the form
\begin{align}
\label{gengrad}
\nabla \mathcal{L}(\vct{w}):=\frac{2}{n}\sum_{i=1}^n\left(\text{ReLU}\left(\langle \vct{w},\vct{x}_i\rangle\right)-y_i\right)\left(1+\sgn{\langle\vct{w},\vct{x}_i\rangle}\right)\vct{x}_i.
\end{align}
Therefore, projected gradient descent takes the form
\begin{align}
\label{iters}
\vct{w}_{\tau+1}=\mathcal{P}_{\mathcal{K}}\left(\vct{w}_\tau-\mu_\tau\nabla \mathcal{L}(\vct{w}_\tau)\right),
\end{align}
where $\mu_\tau$ is the step size and $\mathcal{K}=\{\vct{w}\in\R^d: \mathcal{R}(\vct{w})\le R\}$ is the constraint set with $\mathcal{P}_{\mathcal{K}}$ denoting the Euclidean projection onto this set.
\begin{theorem}\label{Athm}
Let $\vct{w}^*\in\R^d$ be an arbitrary weight vector and $\mathcal{R}:\R^d\rightarrow\R$ be a proper function (convex or nonconvex). Suppose the feature vectors $\vct{x}_i\in\R^{d}$ are i.i.d.~Gaussian random vectors distributed as $\mathcal{N}(\vct{0},\mtx{I})$ with the corresponding labels given by
\begin{align*}
\vct{y}_i=\max\left(0,\langle\vct{x}_i,\vct{w}^*\rangle\right).
\end{align*} 
To estimate $\vct{w}^*$, we start from the initial point $\vct{w}_0=\vct{0}$ and apply the Projected Gradient  (PGD) updates of the form
\begin{align}
\label{myrealupdateA2}
\vct{w}_{\tau+1}=\mathcal{P}_{\mathcal{K}}\left(\vct{w}_\tau-\mu_\tau\nabla \mathcal{L}(\vct{w}_\tau)\right),
\end{align}
with $\mathcal{K}:=\{\vct{w}\in\R^d:\text{ }\mathcal{R}(\vct{w})\le \mathcal{R}(\vct{w}^*)\}$ and $\nabla \mathcal{L}$ defined via \eqref{gengrad}. Also set the learning parameter sequence $\mu_\tau=1$ for all $\tau=0,1,2,\ldots$ and let $n_0=\mathcal{M}(\mathcal{R},\vct{w}^*)$, defined by \ref{PTcurve}, be our lower bound on the number of measurements. Also assume
\begin{align}
\label{nummeaslin}
n>c n_0,
\end{align}
holds for a fixed numerical constant $c$. Then there is an event of probability at least $1-9e^{-\gamma n}$ such that on this event the updates \eqref{myrealupdateA2} obey
\begin{align}
\label{ratemin}
\twonorm{\vct{w}_\tau-\vct{w}^*}\le\left(\frac{1}{2}\right)^{\tau}\twonorm{\vct{w}^*}.
\end{align}
Here $\gamma$ is a fixed numerical constant.
\end{theorem}
The first interesting and perhaps surprising aspect of this result is its generality: it applies not only to convex regularization functions but also nonconvex ones! As we mentioned earlier the optimization problem in \eqref{opt} is not known to be tractable even for convex regularizers.  Despite the nonconvexity of both the objective and regularizer, the theorem above shows that with a near minimal number of data samples, projected gradient descent provably learns the original weight vector $\vct{w}^*$ without getting trapped in any local optima.

Another interesting aspect of the above result is that the convergence rate is linear. Therefore, to achieve a relative error of $\epsilon$ the total number of iterations is on the order of $\mathcal{O}(\log(1/\epsilon))$. Thus the overall computational complexity is on the order of $\mathcal{O}\left(nd\log(1/\epsilon)\right)$ (in general the cost is the total number of iterations multiplied by the cost of applying the feature matrix $\mtx{X}$ and its transpose). As a result, the computational complexity is also now optimal in terms of dependence on the matrix dimensions. Indeed, for a dense matrix even verifying that a good solution has been achieved requires one matrix-vector multiplication which takes $\mathcal{O}(nd)$ time.

\section{Discussions and prior art}\label{PART}
\label{mysec}
There is a large body of work on learning nonlinear models. A particular class of such problems that have been studied are the so called idealized Single Index Models (SIMs) \cite{horowitz1996direct, ichimura1993semiparametric}. In these problems the inputs are labeled examples $\{(\vct{x}_i,y_i)\}_{i=1}^n\in\R^d\times \R$ which are guaranteed to satisfy $y_i=f(\langle\vct{w},\vct{x}_i\rangle)$ for some $\vct{w}\in\R^d$ and nondecreasing (Lipchitz continuous) $f:\R\rightarrow \R$. The goal in this problem is to find a (nearly) accurate such $f$ and $\vct{w}$. An interesting polynomial-time algorithm called the Isotron exists for this problem \cite{kalai2009isotron, kakade2011efficient}. In principle, this approach can also be used to fit ReLUs. However, these results differ from ours in term of both assumptions and results. On the one had, the assumptions are slightly more restrictive as they require bounded features $\vct{x}_i$, outputs $y_i$ and weights. On the other hand, these result hold for much more general distributions and more general models than the realizable model studied in this paper. These results also do not apply in the high dimensional regime where the number of observations is significantly smaller than the number of parameters (see \cite{ganti2015learning} for some results in this direction). In the realizable case, the Isotron result require $\mathcal{O}(\frac{1}{\epsilon})$ iterations to achieve $\epsilon$ error in objective value. In comparison, our results guarantee convergence to a solution with relative error $\epsilon$ ($\twonorm{\vct{w}_\tau-\vct{w}^*}/\twonorm{\vct{w}^*}\le \epsilon$) after $\log\left(1/\epsilon\right)$ iterations. Focusing on the specific case of ReLU functions, an interesting recent result \cite{goel2016reliably} shows that reliable learning of ReLUs is possible under very general but bounded distributional assumptions. To achieve an accuracy of $\epsilon$ the algorithm runs in poly$(1/\epsilon)$ time. In comparison, as mentioned earlier our result rquires $\log(1/\epsilon)$ iterations for reliable parameter estimation. We note however we study the problem in different settings and a direct comparison is not possible between the two results.


\section{Proofs}
\subsection{Preliminaries}
In this section we gather some useful results on concentration of stochastic processes which will be crucial in our proofs. These results are mostly adapted from \cite{soltanolkotabi2017structured, WF, soltanolkotabi2014algorithms}. 
We begin with a lemma which is a direct consequence of Gordon's escape from the mesh lemma \cite{Gor}.
\begin{lemma}\label{gordontype}Assume $\mathcal{C}\subset\R^d$ is a cone and $\mathbb{S}^{d-1}$ is the unit sphere of $\R^d$. Also assume that
\begin{align*}
n\ge \max\left(20\frac{\omega^2(\mathcal{C}\cap\mathbb{S}^{d-1})}{\delta^2},\frac{1}{2\delta}-1\right),
\end{align*}
for a fixed numerical constant $c$. Then for all $\vct{h}\in\mathcal{C}$
\begin{align*}
\abs{\frac{1}{n}\sum_{i=1}^n(\langle\vct{x}_i,\vct{h}\rangle)^2-\twonorm{\vct{h}}^2}\le \delta\twonorm{\vct{h}}^2,
\end{align*}
holds with probability at least $1-2e^{-\frac{\delta^2}{360}n}$. 
\end{lemma}
We also need a generalization of the above lemma stated below.
\begin{lemma}[\cite{soltanolkotabi2017structured}]\label{gordontypenonsym}Assume $\mathcal{C}\subset\R^d$ is a cone (not necessarily convex) and $\mathbb{S}^{d-1}$ is the unit sphere of $\R^d$. Also assume that
\begin{align*}
n\ge \max\left(80\frac{\omega^2(\mathcal{C}\cap\mathbb{S}^{d-1})}{\delta^2},\frac{2}{\delta}-1\right),
\end{align*}
for a fixed numerical constant $c$. Then for all $\vct{u},\vct{h}\in\mathcal{C}$
\begin{align*}
\abs{\frac{1}{n}\sum_{i=1}^n\langle\vct{x}_i,\vct{u}\rangle\langle\vct{x}_i,\vct{h}\rangle-\vct{u}^*\vct{h}}\le \delta\twonorm{\vct{u}}\twonorm{\vct{h}},
\end{align*}
holds with probability at least $1-6e^{-\frac{\delta^2}{1440}n}$. 
\end{lemma}
We next state a generalization of Gordon's escape through the mesh lemma also from \cite{soltanolkotabi2017structured}.
\begin{lemma}[\cite{soltanolkotabi2017structured}]\label{GTtypelem} Let $\vct{s}\in\R^d$ be fixed vector with nonzero entries and construct the diagonal matrix $\mtx{S}=\text{diag}(\vct{s})$. Also, let $\mtx{X}\in\R^{n\times d}$ have i.i.d.~$\mathcal{N}(0,1)$ entries. Furthermore, assume $\mathcal{T}\subset\R^d$ and define
\begin{align*}
b_d(\vct{s})=\E[\twonorm{\mtx{S}\vct{g}}],
\end{align*}
where $\vct{g}\in\R^d$ is distributed as $\mathcal{N}(\vct{0},\mtx{I}_n)$. Define
\begin{align*}
\sigma(\mathcal{T}):=\underset{\vct{v}\in\mathcal{T}}{\max}\text{ }\twonorm{\vct{v}},
\end{align*}
then for all $\vct{u}\in\mathcal{T}$ 
\begin{align*}
\abs{\twonorm{\mtx{S}\mtx{A}\vct{u}}-b_d(\vct{s})\twonorm{\vct{u}}}\le \infnorm{\vct{s}}\omega(\mathcal{T})+\eta,
\end{align*}
holds with probability at least 
\begin{align*}
1-6e^{-\frac{\eta^2}{8\infnorm{\vct{s}}^2\sigma^2(\mathcal{T})}}.
\end{align*}
\end{lemma}
The previous lemma leads to the following Corollary. 
\begin{corollary}\label{coroG}
Let $\vct{s}\in\R^d$ be fixed vector with nonzero entries and assume $\mathcal{T}\subset\mathcal{B}^d$. Furthermore, assume
\begin{align*}
\twonorm{\vct{s}}^2\ge \max\left(20\infnorm{\vct{s}}^2\frac{\omega^2(\mathcal{T})}{\delta^2},\frac{3}{2\delta}-1\right).
\end{align*}
Then for all $\vct{u}\in\mathcal{T}$, 
\begin{align*}
\abs{\frac{\sum_{i=1}^ns_i^2(\langle\vct{x}_i,\vct{u}\rangle)^2}{\twonorm{\vct{s}}^2}-\twonorm{\vct{u}}^2}\le \delta,
\end{align*}
holds with probability at least $1-6e^{-\frac{\delta^2}{1440}\twonorm{\vct{s}}^2}$.
\end{corollary}
\subsection{Convergence proof (Proof of Theorem \ref{Athm})}
In this section we shall prove Theorem \ref{Athm}. Throughout, we use the shorthand $\mathcal{C}$ to denote the descent cone of $\mathcal{R}$ at $\vct{w}^*$, i.e.~$\mathcal{C}=\mathcal{C}_{\mathcal{R}}(\vct{w}^*)$. We begin by analyzing the first iteration. Using $\vct{w}_0=\vct{0}$ we have
\begin{align*}
\vct{w}_1:=\mathcal{P}_{\mathcal{K}}\left(\vct{w}_0-\mu_0\nabla \mathcal{L}(\vct{w}_0)\right)=\mathcal{P}_{\mathcal{K}}\left(\frac{2}{n}\sum_{i=1}^n y_i\vct{x}_i\right)=\mathcal{P}_{\mathcal{K}}\left(\frac{2}{n}\sum_{i=1}^n \text{ReLU}(\langle \vct{x}_i,\vct{w}^*\rangle)\vct{x}_i\right).
\end{align*}
We use the argument of \cite{soltanolkotabi2017structured}[Page 25, inequality (7.34)] which shows that
\begin{align}
\label{myeq11}
\tn{\des_{1}-\vct{w}^*}\leq 2\cdot\sup_{\vct{u}\in\Cc\cap\Bc^{d}} \vct{u}^T\left(\frac{2}{n}\sum_{i=1}^n \text{ReLU}(\langle \vct{x}_i,\vct{w}^*\rangle)\vct{x}_i-\vct{w}^*\right).
\end{align}
Using ReLU$(z)=\frac{z+\abs{z}}{2}$ we have
\begin{align}
\label{eqrel}
\frac{2}{n}\sum_{i=1}^n \text{ReLU}(\langle \vct{x}_i,\vct{w}^*\rangle)\langle\vct{x}_i,\vct{u}\rangle-\langle\vct{u},\vct{w}^*\rangle=\vct{u}^T\left(\frac{1}{n}\mtx{X}^T\mtx{X}-\mtx{I}\right)\vct{w}^*+\frac{1}{n}\sum_{i=1}^n \abs{\langle \vct{x}_i,\vct{w}^*\rangle}\langle\vct{x}_i,\vct{u}\rangle.
\end{align}
We proceed by bounding the first term in the above equality. To this aim we decompose $\vct{u}$ in the direction parallel/perpendicular to that of $\vct{w}^*$ and arrive at
\begin{align}
\label{tmp331}
\vct{u}^T\left(\frac{1}{n}\mtx{X}^T\mtx{X}-\mtx{I}\right)\vct{w}^*=&\frac{(\vct{u}^T\vct{w}^*)}{\twonorm{\vct{w}^*}^2}\left(\vct{w}^*\right)^T\left(\frac{1}{n}\mtx{X}^T\mtx{X}-\mtx{I}\right)\vct{w}^*+\frac{1}{n}\Bigg\langle\mtx{X}\left(\mtx{I}-\frac{\vct{w}^*\left(\vct{w}^*\right)^T}{\twonorm{\vct{w}^*}^2}\right)\vct{u},\mtx{X}\vct{w}^*\Bigg\rangle,\nonumber\\
\sim&(\vct{u}^T\vct{w}^*)\left(\frac{\twonorm{\vct{g}}^2}{n}-1\right)+\frac{\twonorm{\vct{w}^*}}{\sqrt{n}}\vct{a}^T\left(\mtx{I}-\frac{\vct{w}^*\left(\vct{w}^*\right)^T}{\twonorm{\vct{w}^*}^2}\right)\vct{u},\nonumber\\
\le&\twonorm{\vct{w}^*}\abs{\frac{\twonorm{\vct{g}}^2}{n}-1}+\frac{\twonorm{\vct{w}^*}}{\sqrt{n}}\underset{\vct{u}\in\mathcal{C}\cap\mathcal{B}^d}{\sup}\text{ }\vct{a}^T\left(\mtx{I}-\frac{\vct{w}^*\left(\vct{w}^*\right)^T}{\twonorm{\vct{w}^*}^2}\right)\vct{u},
\end{align}
with $\vct{g}\in\R^n$ and $\vct{a}\in\R^d$ are independent random Gaussian random vectors distributed as $\mathcal{N}(\vct{0},\mtx{I}_d)$ and $\mathcal{N}(\vct{0},\mtx{I}_n)$. By concentration of Chi-squared random variables
\begin{align}
\label{tmp3311}
\abs{\frac{\twonorm{\vct{g}}^2}{n}-1}\le \Delta,
\end{align}
holds with probability at least $1-2e^{-n\frac{\Delta^2}{8}}$. Also,
\begin{align}
\label{tmp3312}
\frac{1}{\sqrt{n}}\vct{a}^T\left(\mtx{I}-\frac{\vct{w}^*\left(\vct{w}^*\right)^T}{\twonorm{\vct{w}^*}^2}\right)\vct{u}\le \frac{1}{\sqrt{n}}\left(\omega\left(\mathcal{C}\cap\mathcal{B}^{d}\right)+\eta\right),
\end{align}
holds with probability at least $1-e^{-\frac{\eta^2}{2}}$. Plugging \eqref{tmp3311} with $\Delta=\frac{\delta}{6}$ and \eqref{tmp3312}  with $\eta=\frac{\delta}{6}\sqrt{n}$ into \eqref{tmp331}, as long as
\begin{align*}
n\ge \frac{36}{\delta^2}\omega^2\left(\mathcal{C}\cap\mathcal{B}^{d}\right),
\end{align*}
then
\begin{align}
\label{tmp351}
\sup_{\vct{u}\in\Cc\cap\Bc^{d}}\text{ }\vct{u}^T\left(\frac{1}{n}\mtx{X}^T\mtx{X}-\mtx{I}\right)\vct{w}^*\le \frac{\delta}{2} \twonorm{\vct{w}^*},
\end{align}
holds with probability at least $1-3e^{-n\frac{\delta^2}{288}}$. 

We now focus on bounding the second term in \eqref{eqrel}. To this aim we decompose $\vct{u}$ in the direction parallel/perpendicular to that of $\vct{w}^*$ and arrive at
\begin{align}
\label{ineq2}
\abs{\frac{1}{n}\sum_{i=1}^n \abs{\langle \vct{x}_i,\vct{w}^*\rangle}\langle\vct{x}_i,\vct{u}\rangle}=&\abs{(\vct{u}^T\vct{w}^*)\frac{1}{n}\sum_{i=1}^n \frac{\abs{\langle \vct{x}_i,\vct{w}^*\rangle}\langle\vct{x}_i,\vct{w}^*\rangle}{\twonorm{\vct{w}^*}^2}+\frac{1}{n}\sum_{i=1}^n \abs{\langle \vct{x}_i,\vct{w}^*\rangle}\langle\vct{x}_i,\vct{u}_{\perp}\rangle},\nonumber\\
\le&\twonorm{\vct{w}^*}\abs{\frac{1}{n}\sum_{i=1}^n \frac{\abs{\langle \vct{x}_i,\vct{w}^*\rangle}\langle\vct{x}_i,\vct{w}^*\rangle}{\twonorm{\vct{w}^*}^2}}+\abs{\frac{1}{n}\sum_{i=1}^n \abs{\langle \vct{x}_i,\vct{w}^*\rangle}\langle\vct{x}_i,\vct{u}_{\perp}\rangle}.
\end{align}
with $\vct{u}_\perp=\left(\mtx{I}-\frac{\vct{w}^*(\vct{w}^*)^T}{\twonorm{\vct{w}^*}^2}\right)\vct{u}$. Now note that $\frac{\abs{\langle \vct{x}_i,\vct{w}^*\rangle}\langle\vct{x}_i,\vct{w}^*\rangle}{\twonorm{\vct{w}^*}^2}$ is sub-exponential with norm bounded by
\begin{align*}
\bigg\|\frac{\abs{\langle \vct{x}_i,\vct{w}^*\rangle}\langle\vct{x}_i,\vct{w}^*\rangle}{\twonorm{\vct{w}^*}^2}\bigg\|_{\psi_1}\le c,  
\end{align*}
with fixed numerical constant. Thus by Bernstein's type inequality (\cite{vershynin2010introduction}[Proposition 5.16])
\begin{align}
\label{tmp221}
\abs{\frac{1}{n}\sum_{i=1}^n \frac{\abs{\langle \vct{x}_i,\vct{w}^*\rangle}\langle\vct{x}_i,\vct{w}^*\rangle}{\twonorm{\vct{w}^*}^2}}\le t,
\end{align}
holds with probability at least $1-2e^{-\gamma n\min\left(t^2,t\right)}$ with $\gamma$ a fixed numerical constant. Also note that
\begin{align*}
\frac{1}{n}\sum_{i=1}^n \abs{\langle \vct{x}_i,\vct{w}^*\rangle}\langle\vct{x}_i,\vct{u}_{\perp}\rangle\sim \sqrt{\frac{1}{n}\sum_{i=1}^n\abs{\langle \vct{x}_i,\vct{w}^*\rangle}^2}\frac{1}{\sqrt{n}}\langle \vct{g},\vct{u}_\perp\rangle.
\end{align*}
Furthermore,
\begin{align*}
\frac{1}{n}\sum_{i=1}^n\abs{\langle \vct{x}_i,\vct{w}^*\rangle}^2\le(1+\Delta)\twonorm{\vct{w}^*}^2,
\end{align*}
holds with probability at least $1-2e^{-n\frac{\Delta^2}{8}}$ and
\begin{align*}
\underset{\vct{u}\in\mathcal{C}\cap\mathbb{S}^{d-1}}{\sup}\text{ }\frac{1}{\sqrt{n}}\abs{\langle\vct{g},\vct{u}_{\perp}\rangle}\le \frac{(2\omega\left(\mathcal{C}\cap\mathbb{S}^{d-1}\right)+\eta)}{\sqrt{n}},
\end{align*}
holds with probability at least $1-e^{-\frac{\eta^2}{2}}$. Combining the last two inequalities we conclude that
\begin{align}
\label{tmp222}
\abs{\frac{1}{n}\sum_{i=1}^n \abs{\langle \vct{x}_i,\vct{w}^*\rangle}\langle\vct{x}_i,\vct{u}_{\perp}\rangle}\le\sqrt{1+\Delta}\frac{(2\omega\left(\mathcal{C}\cap\mathbb{S}^{d-1}\right)+\eta)}{\sqrt{n}}\twonorm{\vct{w}^*},
\end{align} 
holds with probability at least $1-2e^{-n\frac{\Delta^2}{8}}-e^{-\frac{\eta^2}{2}}$. Plugging \eqref{tmp221} and \eqref{tmp222} with $t=\frac{\delta}{6}$, $\Delta=1$, and $\eta=\frac{\delta}{6\sqrt{2}}\sqrt{n}$ into \eqref{ineq2}
\begin{align}
\label{mymyeq}
\abs{\frac{1}{n}\sum_{i=1}^n \abs{\langle \vct{x}_i,\vct{w}^*\rangle}\langle\vct{x}_i,\vct{u}\rangle}\le\frac{\delta}{2}\twonorm{\vct{w}^*},
\end{align}
holds with probability at least $1-3e^{-\gamma n\delta^2}-2e^{-\frac{n}{8}}$ as long as
\begin{align*}
n\ge 288\frac{\omega^2\left(\mathcal{C}\cap\mathbb{S}^{d-1}\right)}{\delta^2}.
\end{align*}
Thus pluggin \eqref{tmp351} and \eqref{mymyeq} into \eqref{myeq11} we conclude that for $\delta=7/400$
\begin{align*}
\tn{\des_{1}-\vct{w}^*}\le& 2\cdot\sup_{\vct{u}\in\Cc\cap\Bc^{d}} \vct{u}^T\left(\frac{2}{n}\sum_{i=1}^n \text{ReLU}(\langle \vct{x}_i,\vct{w}^*\rangle)\vct{x}_i-\vct{w}^*\right),\\
\le& 2\delta\twonorm{\vct{w}^*},\\
\le&\frac{7}{200} \twonorm{\vct{w}^*},
\end{align*}
holds with probability at least $1-8e^{-\gamma n}$ as long as
\begin{align*}
n\ge c\omega^2\left(\mathcal{C}\cap\mathbb{S}^{d-1}\right),
\end{align*}
for a fixed numerical constant $c$.

To introduce our general convergence analysis we begin by defining 
\begin{align*}
E(\epsilon)=\big\{\vct{w}\in\R^d:\mathcal{R}(\vct{w})\le \mathcal{R}(\vct{w}^*),\text{ }\twonorm{\vct{w}-\vct{w}^*}\le\epsilon\twonorm{\vct{w}^*}\big\}\quad\text{with}\quad\epsilon=\frac{7}{200}.
\end{align*}
To prove Theorem \ref{Athm} we use the argument of \cite{soltanolkotabi2017structured}[Page 25, inequality (7.34)] which shows that if we apply the projected gradient descent update
\begin{align*}
\vct{w}_{\tau+1}=\mathcal{P}_{\mathcal{K}}\left(\vct{w}_\tau-\nabla\mathcal{L}(\vct{w}_\tau)\right),
\end{align*}
the error $\vct{h}_\tau=\vct{w}_\tau-\vct{w}^*$ obeys
\begin{align}
\label{interpfthm12}
\tn{\vct{h}_{\tau+1}}=\tn{\des_{\tau+1}-\vct{w}^*}\leq 2\cdot\sup_{\vct{u}\in\Cc\cap\Bc^{n}} \vct{u}^*\left(\h_{\tau}-\nabla\mathcal{L}(\vct{w}_\tau)\right).
\end{align}
To complete the convergence analysis it is then sufficient to prove
\begin{align}
\label{mainintemp}
\sup_{\vct{u}\in\Cc\cap\Bc^{n}} \vct{u}^*\left(\h_{\tau}-\nabla\mathcal{L}(\vct{w}_\tau)\right)\le\frac{1}{4}\twonorm{\h_{\tau}}=\frac{1}{4}\twonorm{\vct{w}_\tau-\vct{w}^*}.
\end{align}
We will instead prove that the following stronger result holds for all $\vct{u}\in\Cc\cap\Bc^{n}$ and $\vct{w}\in E(\epsilon)$
\begin{align}
\label{mainineq}
\vct{u}^*\left(\vct{w}-\vct{w}^*-\nabla\mathcal{L}(\vct{w})\right)\le\frac{1}{4}\twonorm{\vct{w}-\vct{w}^*}.
\end{align}
The equation \eqref{mainineq} above implies \eqref{mainintemp} which when combined with \eqref{interpfthm12} proves the convergence result of the Theorem (specifically equation \eqref{ratemin}).

The rest of this section is dedicated to proving \eqref{mainineq}. To this aim note that $\text{ReLU}(\langle \vct{x}_i,\vct{w}\rangle)=\frac{\langle\vct{x}_i,\vct{w}\rangle+\abs{\langle\vct{x}_i,\vct{w}\rangle}}{2}$. Therefore, the loss function can alternatively be written as
\begin{align*}
\mathcal{L}(\vct{w})=\frac{1}{4n}\sum_{i=1}^n\left(\abs{\langle \vct{x}_i,\vct{w}\rangle}-\abs{\langle\vct{x}_i,\vct{w}^*\rangle}\right)^2+\frac{1}{4n}\sum_{i=1}^n\left(\langle \vct{x}_i,\vct{w}-\vct{w}^*\rangle\right)^2+\frac{1}{2n}\sum_{i=1}^n \left(\abs{\langle \vct{x}_i,\vct{w}\rangle}-\abs{\langle\vct{x}_i,\vct{w}^*\rangle}\right)\left(\langle \vct{x}_i,\vct{w}-\vct{w}^*\rangle\right).
\end{align*}
Thus
\begin{align*}
\langle\nabla \mathcal{L}(\vct{w}), \vct{u}\rangle =&\frac{1}{2n}\sum_{i=1}^n\left(\langle \vct{x}_i,\vct{w}\rangle-\sgn{\langle \vct{x}_i,\vct{w}\rangle}\abs{\langle\vct{x}_i,\vct{w}^*\rangle}\right)\langle\vct{x}_i,\vct{u}\rangle\\
&+\frac{1}{2n}\sum_{i=1}^n\langle \vct{x}_i,\vct{w}-\vct{w}^*\rangle\langle\vct{x}_i,\vct{u}\rangle+\frac{1}{2n}\sum_{i=1}^n \left(\abs{\langle \vct{x}_i,\vct{w}\rangle}-\abs{\langle\vct{x}_i,\vct{w}^*\rangle}\right)\langle \vct{x}_i,\vct{u}\rangle\\
&+\frac{1}{2n}\sum_{i=1}^n \sgn{\langle \vct{x}_i,\vct{w}\rangle}\langle \vct{x}_i,\vct{w}-\vct{w}^*\rangle\langle \vct{x}_i,\vct{u}\rangle\\
=&\frac{1}{n}\sum_{i=1}^n\langle \vct{x}_i,\vct{w}-\vct{w}^*\rangle\langle \vct{x}_i,\vct{u}\rangle+\frac{1}{n}\sum_{i=1}^n \left(\abs{\langle \vct{x}_i,\vct{w}\rangle}-\abs{\langle\vct{x}_i,\vct{w}^*\rangle}\right)\langle \vct{x}_i,\vct{u}\rangle\\
&+\frac{1}{2n}\sum_{i=1}^n\left(1-\sgn{\langle\vct{x}_i,\vct{w}^*\rangle}\sgn{\langle\vct{x}_i,\vct{w}\rangle}+\sgn{\langle\vct{x}_i,\vct{w}^*\rangle}-\sgn{\langle\vct{x}_i,\vct{w}\rangle}\right)\langle\vct{x}_i,\vct{w}^*\rangle\langle\vct{x}_i,\vct{u}\rangle\\
=&\frac{1}{n}\sum_{i=1}^n\langle \vct{x}_i,\vct{w}-\vct{w}^*\rangle\langle \vct{x}_i,\vct{u}\rangle+\frac{1}{n}\sum_{i=1}^n \sgn{\langle \vct{x}_i,\vct{w}^*\rangle}\langle \vct{x}_i,\vct{w}-\vct{w}^*\rangle\langle\vct{x}_i,\vct{u}\rangle\\
&+\frac{1}{n}\sum_{i=1}^n \left(\sgn{\langle \vct{x}_i,\vct{w}\rangle}-\sgn{\langle \vct{x}_i,\vct{w}^*\rangle}\right)\langle \vct{x}_i,\vct{w}-\vct{w}^*\rangle\langle\vct{x}_i,\vct{u}\rangle\\
&+\frac{1}{2n}\sum_{i=1}^n\left(1-\sgn{\langle\vct{x}_i,\vct{w}^*\rangle}\sgn{\langle\vct{x}_i,\vct{w}\rangle}-\sgn{\langle\vct{x}_i,\vct{w}^*\rangle}+\sgn{\langle\vct{x}_i,\vct{w}\rangle}\right)\langle\vct{x}_i,\vct{w}^*\rangle\langle\vct{x}_i,\vct{u}\rangle\\
=&\frac{1}{n}\sum_{i=1}^n\langle \vct{x}_i,\vct{w}-\vct{w}^*\rangle\langle \vct{x}_i,\vct{u}\rangle+\frac{1}{n}\sum_{i=1}^n \sgn{\langle \vct{x}_i,\vct{w}^*\rangle}\langle \vct{x}_i,\vct{w}-\vct{w}^*\rangle\langle\vct{x}_i,\vct{u}\rangle\\
&+\frac{1}{n}\sum_{i=1}^n \left(\sgn{\langle \vct{x}_i,\vct{w}\rangle}-\sgn{\langle \vct{x}_i,\vct{w}^*\rangle}\right)\langle \vct{x}_i,\vct{w}-\vct{w}^*\rangle\langle\vct{x}_i,\vct{u}\rangle\\
&+\frac{1}{2n}\sum_{i=1}^n\left(1-\sgn{\langle\vct{x}_i,\vct{w}^*\rangle}\right)\left(\sgn{\langle\vct{x}_i,\vct{w}^*\rangle}-\sgn{\langle\vct{x}_i,\vct{w}\rangle}\right)\abs{\langle\vct{x}_i,\vct{w}^*\rangle}\langle\vct{x}_i,\vct{u}\rangle
\end{align*}
Now defining $\vct{h}=\vct{w}-\vct{w}^*$ we conclude that
\begin{align*}
\langle \vct{u} ,\vct{w}-\vct{w}^*-\nabla \mathcal{L}(\vct{w})\rangle=&\langle \vct{u} ,\vct{h}-\nabla \mathcal{L}(\vct{w})\rangle,\nonumber\\
=&\vct{u}^T\left(\mtx{I}-\frac{1}{n}\mtx{X}\mtx{X}^T\right)\vct{h}-\frac{1}{n}\sum_{i=1}^n \sgn{\langle \vct{x}_i,\vct{w}^*\rangle}\langle \vct{x}_i,\vct{h}\rangle\langle\vct{x}_i,\vct{u}\rangle,\nonumber\\
&+\frac{1}{n}\sum_{i=1}^n \left(1-\sgn{\langle \vct{x}_i,\vct{w}\rangle}\sgn{\langle \vct{x}_i,\vct{w}^*\rangle}\right)\sgn{\langle \vct{x}_i,\vct{w}^*\rangle}\langle \vct{x}_i,\vct{h}\rangle\langle\vct{x}_i,\vct{u}\rangle,\nonumber\\
&+\frac{1}{2n}\sum_{i=1}^n\sgn{\langle\vct{x}_i,\vct{w}\rangle}\left(1-\sgn{\langle\vct{x}_i,\vct{w}^*\rangle}\right)\left(1-\sgn{\langle\vct{x}_i,\vct{w}\rangle}\sgn{\langle\vct{x}_i,\vct{w}^*\rangle}\right)\abs{\langle\vct{x}_i,\vct{w}^*\rangle}\langle\vct{x}_i,\vct{u}\rangle.
\end{align*}
Now define $\vct{h}_{\perp}=\vct{h}-\frac{(\vct{h}^T\vct{w}^*)}{\twonorm{\vct{w}^*}^2}\vct{w}^*$. Using this we can rewrite the previous expression in the form
\begin{align}
\langle \vct{u} ,\vct{w}-\vct{w}^*-\nabla \mathcal{L}(\vct{w})\rangle=&\vct{u}^T\left(\mtx{I}-\frac{1}{n}\mtx{X}\mtx{X}^T\right)\vct{h}-\frac{1}{n}\sum_{i=1}^n \sgn{\langle \vct{x}_i,\vct{w}^*\rangle}\langle \vct{x}_i,\vct{h}\rangle\langle\vct{x}_i,\vct{u}\rangle,\nonumber\\
&+\frac{1}{n}\sum_{i=1}^n \left(1-\sgn{\langle \vct{x}_i,\vct{w}\rangle}\sgn{\langle \vct{x}_i,\vct{w}^*\rangle}\right)\sgn{\langle \vct{x}_i,\vct{w}^*\rangle}\langle \vct{x}_i,\vct{h}_{\perp}\rangle\langle\vct{x}_i,\vct{u}\rangle,\nonumber\\
&+\frac{\langle\vct{h},\vct{w}^*\rangle}{\twonorm{\vct{w}^*}^2}\frac{1}{n}\sum_{i=1}^n \left(1-\sgn{\langle \vct{x}_i,\vct{w}\rangle}\sgn{\langle \vct{x}_i,\vct{w}^*\rangle}\right)\abs{\langle \vct{x}_i,\vct{w}^*\rangle}\langle\vct{x}_i,\vct{u}\rangle,\nonumber\\
&+\frac{1}{2n}\sum_{i=1}^n\sgn{\langle\vct{x}_i,\vct{w}\rangle}\left(1-\sgn{\langle\vct{x}_i,\vct{w}^*\rangle}\right)\left(1-\sgn{\langle\vct{x}_i,\vct{w}\rangle}\sgn{\langle\vct{x}_i,\vct{w}^*\rangle}\right)\abs{\langle\vct{x}_i,\vct{w}^*\rangle}\langle\vct{x}_i,\vct{u}\rangle\nonumber\\
=&\vct{u}^T\left(\mtx{I}-\frac{1}{n}\mtx{X}\mtx{X}^T\right)\vct{h}-\frac{1}{n}\sum_{i=1}^n \sgn{\langle \vct{x}_i,\vct{w}^*\rangle}\langle \vct{x}_i,\vct{h}\rangle\langle\vct{x}_i,\vct{u}\rangle,\nonumber\\
&+\frac{1}{n}\sum_{i=1}^n \left(1-\sgn{\langle \vct{x}_i,\vct{w}\rangle}\sgn{\langle \vct{x}_i,\vct{w}^*\rangle}\right)\sgn{\langle \vct{x}_i,\vct{w}^*\rangle}\langle \vct{x}_i,\vct{h}_{\perp}\rangle\langle\vct{x}_i,\vct{u}\rangle,\nonumber\\
&+\frac{1}{n}\sum_{i=1}^n\bigg[\frac{\sgn{\langle\vct{x}_i,\vct{w}\rangle}}{2}\left(1-\sgn{\langle\vct{x}_i,\vct{w}^*\rangle}\right)+\frac{\langle\vct{h},\vct{w}^*\rangle}{\twonorm{\vct{w}^*}^2}\bigg]\nonumber\\
&\quad\quad\left(1-\sgn{\langle\vct{x}_i,\vct{w}\rangle}\sgn{\langle\vct{x}_i,\vct{w}^*\rangle}\right)\abs{\langle\vct{x}_i,\vct{w}^*\rangle}\langle\vct{x}_i,\vct{u}\rangle
\label{mymain}
\end{align}
We now proceed by bounding each of the four terms in \eqref{mymain} and then combine them in Section \ref{all4}.
\subsubsection{Bounding the first term in \eqref{mymain}}
To bound the first term we use Lemma \ref{gordontypenonsym}, which implies that as long as 
\begin{align*}
n\ge \max\left(80\frac{n_0}{\delta^2},\frac{2}{\delta}-1\right),
\end{align*}
then for all $\vct{u}\in\Cc\cap\Bc^{n}$ and $\vct{h}\in E(\epsilon)$
\begin{align}
\label{myinter1}
\vct{u}^*\left(\mtx{I}-\frac{1}{n}\mtx{X}^*\mtx{X}\right)\vct{h}\le\delta \twonorm{\vct{h}},
\end{align}
holds with probability at least $1-6e^{-\frac{\delta^2}{1440}n}$.
\subsubsection{Bounding the second term in \eqref{mymain}}
To bound the second term we first define
\begin{align*}
\vct{u}_{\perp}=\vct{u}-\frac{(\vct{u}^T\vct{w}^*)}{\twonorm{\vct{w}^*}^2}\vct{w}^*\quad\text{and}\quad\vct{h}_{\perp}=\vct{h}-\frac{(\vct{h}^T\vct{w}^*)}{\twonorm{\vct{w}^*}^2}\vct{w}^*.
\end{align*}
Now note that
\begin{align}
\label{sectmp}
-\frac{1}{n}\sum_{i=1}^n \sgn{\langle \vct{x}_i,\vct{w}^*\rangle}\langle \vct{x}_i,\vct{h}\rangle\langle\vct{x}_i,\vct{u}\rangle=&-\frac{(\vct{u}^T\vct{w}^*)(\vct{h}^T\vct{w}^*)}{\twonorm{\vct{w}^*}^4}\frac{1}{n}\sum_{i=1}^n\abs{\langle\vct{x}_i,\vct{w}^*\rangle}\langle\vct{x}_i,\vct{w}^*\rangle\nonumber\\
&-\frac{(\vct{u}^T\vct{w}^*)}{\twonorm{\vct{w}^*}^2}\frac{1}{n}\sum_{i=1}^n \abs{\langle \vct{x}_i,\vct{w}^*\rangle}\langle\vct{x}_i,\vct{h}_{\perp}\rangle\nonumber\\
&-\frac{(\vct{h}^T\vct{w}^*)}{\twonorm{\vct{w}^*}^2}\frac{1}{n}\sum_{i=1}^n \abs{\langle \vct{x}_i,\vct{w}^*\rangle}\langle\vct{x}_i,\vct{u}_{\perp}\rangle\nonumber\\
&-\frac{1}{n}\sum_{i=1}^n \sgn{\langle \vct{x}_i,\vct{w}^*\rangle}\langle \vct{x}_i,\vct{h}_{\perp}\rangle\langle\vct{x}_i,\vct{u}_{\perp}\rangle,\nonumber\\
\le&\twonorm{\vct{h}}\abs{\frac{1}{n}\sum_{i=1}^n\frac{\abs{\langle\vct{x}_i,\vct{w}^*\rangle}}{\twonorm{\vct{w}^*}}\frac{\langle\vct{x}_i,\vct{w}^*\rangle}{\twonorm{\vct{w}^*}}}\nonumber\\
&+\abs{\frac{1}{n}\sum_{i=1}^n \frac{\abs{\langle \vct{x}_i,\vct{w}^*\rangle}}{\twonorm{\vct{w}^*}}\langle\vct{x}_i,\vct{h}_{\perp}\rangle}+\twonorm{\vct{h}}\abs{\frac{1}{n}\sum_{i=1}^n \frac{\abs{\langle \vct{x}_i,\vct{w}^*\rangle}}{\twonorm{\vct{w}^*}}\langle\vct{x}_i,\vct{u}_{\perp}\rangle}\nonumber\\
&+\abs{\frac{1}{n}\sum_{i=1}^n \sgn{\langle \vct{x}_i,\vct{w}^*\rangle}\langle \vct{x}_i,\vct{h}_{\perp}\rangle\langle\vct{x}_i,\vct{u}_{\perp}\rangle}.
\end{align}
We now proceed by bound the four terms on the right-hand side of the above inequality. To bound the first term we use \eqref{tmp221} to conclude that
\begin{align}
\label{page111}
\abs{\frac{1}{n}\sum_{i=1}^n \frac{\abs{\langle \vct{x}_i,\vct{w}^*\rangle}\langle\vct{x}_i,\vct{w}^*\rangle}{\twonorm{\vct{w}^*}^2}}\le t,
\end{align}
holds with probability at least $1-2e^{-\gamma n\min\left(t^2,t\right)}$. To bound the second and third terms in \eqref{sectmp} we use \eqref{tmp222} to conclude that 
\begin{align}
\label{page112}
\abs{\frac{1}{n}\sum_{i=1}^n \frac{\abs{\langle \vct{x}_i,\vct{w}^*\rangle}}{\twonorm{\vct{w}^*}}\langle\vct{x}_i,\vct{h}_{\perp}\rangle}\le&\sqrt{1+\Delta}\frac{(2\sqrt{n_0}+\eta\sqrt{n})}{\sqrt{n}}\twonorm{\vct{h}},\nonumber\\
\abs{\frac{1}{n}\sum_{i=1}^n \frac{\abs{\langle \vct{x}_i,\vct{w}^*\rangle}}{\twonorm{\vct{w}^*}}\langle\vct{x}_i,\vct{u}_{\perp}\rangle}\le&\sqrt{1+\Delta}\frac{(2\sqrt{n_0}+\eta\sqrt{n})}{\sqrt{n}},
\end{align}
holds with probability at least $1-2e^{-n\frac{\Delta^2}{8}}-e^{-\frac{\eta^2}{2}n}$. To bound the last term let $\epsilon_i$ be i.i.d.~$\pm 1$ random variables independent from $\vct{x}_i$. Then, 
\begin{align*}
\frac{1}{n}\sum_{i=1}^n \sgn{\langle \vct{x}_i,\vct{w}^*\rangle}\langle \vct{x}_i,\vct{h}_{\perp}\rangle\langle\vct{x}_i,\vct{u}_{\perp}\rangle\sim\frac{1}{n}\sum_{i=1}^n \epsilon_i\langle \vct{x}_i,\vct{h}_{\perp}\rangle\langle\vct{x}_i,\vct{u}_{\perp}\rangle.
\end{align*}
Define $\mathcal{T}= \left(\mtx{I}-\frac{\vct{w}^*(\vct{w}^*)^T}{\twonorm{\vct{w}^*}^2}\right)\mathcal{C}\cap\mathbb{S}^{d-1}$ and note that
\begin{align}
\label{mmeq1}
\abs{\frac{1}{n}\sum_{i=1}^n \epsilon_i\langle \vct{x}_i,\vct{h}_{\perp}\rangle\langle\vct{x}_i,\vct{u}_{\perp}\rangle}\le\left(\underset{\vct{u},\vct{v}\in\mathcal{T}}{\sup}\abs{\frac{1}{n}\sum_{i=1}^n \epsilon_i\langle \vct{x}_i,\vct{u}\rangle\langle\vct{x}_i,\vct{v}\rangle}\right)\twonorm{\vct{h}}.
\end{align}
Now to bound the term in the parenthesis note that by concentration of sums of sub-Gaussian random variables \cite[Proposition 5.10]{vershynin2010introduction}
\begin{align}
\label{mmeq2}
\abs{\frac{1}{n}\sum_{i=1}^n \epsilon_i\langle \vct{x}_i,\vct{u}\rangle\langle\vct{x}_i,\vct{v}\rangle}\le\frac{\Delta}{n} \sqrt{\frac{1}{n}\sum_{i=1}^n (\langle \vct{x}_i,\vct{u}\rangle)^2(\langle\vct{x}_i,\vct{v}\rangle)^2},
\end{align}
holds with probability at least $1-2e^{-\gamma n\Delta^2}$. Now note that 
\begin{align*}
\underset{\vct{u}\in\mathcal{T}}{\sup} \abs{\langle \vct{x}_i, \vct{u}\rangle}\le 2\omega(\mathcal{T})+\eta\le 2\sqrt{n_0}+\eta\sqrt{n},
\end{align*}
holds with probability at least $1-e^{-\frac{\eta^2}{2}n}$. Thus using the union bound
\begin{align*}
\underset{i=1,2,\ldots,n}{\sup}\text{ }\underset{\vct{u}\in\mathcal{T}}{\sup} \abs{\langle \vct{x}_i, \vct{u}\rangle}\le 2\omega(\mathcal{T})+\eta\sqrt{n}\le 2\sqrt{n_0}+\eta\sqrt{n},
\end{align*}
holds with probability at least $1-ne^{-\frac{\eta^2}{2}n}$. Plugging this into \eqref{mmeq2} and using Lemma \ref{GTtypelem} with $\mtx{S}=\mtx{I}$, we conclude that
\begin{align*} 
\underset{\vct{u},\vct{v}\in\mathcal{T}}{\sup}\abs{\frac{1}{n}\sum_{i=1}^n \epsilon_i\langle \vct{x}_i,\vct{u}\rangle\langle\vct{x}_i,\vct{v}\rangle}\le&\frac{\Delta}{n} \left(2\sqrt{n_0}+\eta\sqrt{n}\right)\sqrt{\underset{\vct{v}\in\mathcal{T}}{\sup}\text{ }\frac{1}{n}\sum_{i=1}^n (\vct{x}_i^T\vct{v})^2},\nonumber\\
\le&\frac{\Delta}{n} \left(2\sqrt{n_0}+\eta\sqrt{n}\right)\left(\sqrt{n_0}+(\eta+1)\sqrt{n}\right),
\end{align*}
holds with probability at least $1-2e^{-\gamma n\Delta^2}-ne^{-\frac{\eta^2}{2}n}-6e^{-\frac{\eta^2}{8}n}$. Using the latter inequality in \eqref{mmeq1} we conclude that for all $\vct{h}\in\mathcal{C}$ and $\vct{u}\in\mathcal{C}\cap\mathbb{S}^{n-1}$
\begin{align}
\label{lasttt}
\abs{\frac{1}{n}\sum_{i=1}^n \epsilon_i\langle \vct{x}_i,\vct{h}_{\perp}\rangle\langle\vct{x}_i,\vct{u}_{\perp}\rangle}\le \frac{\Delta}{n} \left(2\sqrt{n_0}+\eta\sqrt{n}\right)\left(\sqrt{n_0}+(\eta+1)\sqrt{n}\right)\twonorm{\vct{h}}
\end{align}
holds with probability at least $1-2e^{-\gamma n\Delta^2}-ne^{-\frac{\eta^2}{2}n}-6e^{-\frac{\eta^2}{8}n}$, completing the bound of the last term of \eqref{sectmp}. Combining \eqref{page111}, \eqref{page112}, and \eqref{lasttt} with $\eta=\Delta=1$, we conclude that
\begin{align}
\label{myinter2}
-\frac{1}{n}\sum_{i=1}^n \sgn{\langle \vct{x}_i,\vct{w}^*\rangle}\langle \vct{x}_i,\vct{h}\rangle\langle\vct{x}_i,\vct{u}\rangle\le\delta\twonorm{\vct{h}},
\end{align}
holds with probability at least $1-2e^{-\gamma \delta^2 n}-(n+10)e^{-\gamma n}$ as long as
\begin{align*}
n\ge c\frac{n_0}{\delta^2}.
\end{align*}
This completes the bound on the second term in \eqref{mymain}. 
\subsubsection{Bounding the third term in \eqref{mymain}}
To bound the third term note that
\begin{align}
\label{third1}
\frac{1}{n}\sum_{i=1}^n \left(1-\sgn{\langle \vct{x}_i,\vct{w}\rangle}\sgn{\langle \vct{x}_i,\vct{w}^*\rangle}\right)&\sgn{\langle \vct{x}_i,\vct{w}^*\rangle}\langle \vct{x}_i,\vct{h}_{\perp}\rangle\langle\vct{x}_i,\vct{u}\rangle\nonumber\\
=&\frac{2}{n}\sum_{i=1}^n\sgn{\langle \vct{x}_i,\vct{w}^*\rangle}\langle \vct{x}_i,\vct{h}_{\perp}\rangle\langle\vct{x}_i,\vct{u}\rangle\mathbb{1}_{\{\langle\vct{x}_i,\vct{w}\rangle\langle\vct{x}_i,\vct{w}^*\rangle\le 0\}},\nonumber\\
\overset{(a)}{\le}& 2\sqrt{\frac{1}{n}\sum_{i=1}^n\abs{\langle \vct{x}_i,\vct{h}_{\perp}\rangle}^2\mathbb{1}_{\{\langle\vct{x}_i,\vct{w}\rangle\langle\vct{x}_i,\vct{w}^*\rangle\le 0\}}}\sqrt{\frac{1}{n}\sum_{i=1}^n\abs{\langle\vct{x}_i,\vct{u}\rangle}^2},
\end{align}
where the last inequality follows from Cauchy Schwarz. Note that by Lemma \ref{gordontype} as long as $n\ge \max\left(80\frac{n_0}{\delta^2},\frac{2}{\delta}-1\right)$, then
\begin{align*}
\frac{1}{n}\sum_{i=1}^n\abs{\langle\vct{x}_i,\vct{u}\rangle}^2\le 1+\delta,
\end{align*}
holds for all $\vct{u}\in\mathcal{C}\cap\mathbb{S}^{d-1}$ with probability at least $1-2e^{-\gamma\delta^2 n}$. Combining the latter inequality with \eqref{third1} we conclude that with high probability
\begin{align}
\label{third2}
\frac{1}{n}\sum_{i=1}^n \left(1-\sgn{\langle \vct{x}_i,\vct{w}\rangle}\sgn{\langle \vct{x}_i,\vct{w}^*\rangle}\right)\sgn{\langle \vct{x}_i,\vct{w}^*\rangle}&\langle \vct{x}_i,\vct{h}_{\perp}\rangle\langle\vct{x}_i,\vct{u}\rangle\nonumber\\
&\le2\sqrt{1+\delta}\sqrt{\frac{1}{n}\sum_{i=1}^n\abs{\langle \vct{x}_i,\vct{h}_{\perp}\rangle}^2\mathbb{1}_{\{\langle\vct{x}_i,\vct{w}\rangle\langle\vct{x}_i,\vct{w}^*\rangle\le 0\}}}.
\end{align}
We now turn our attention to bounding the right-hand side of \eqref{third2}. To this aim first note that as long as $\twonorm{\vct{h}}\le \epsilon\twonorm{\vct{w}^*}$ we have $(\vct{w}^T\vct{w}^*)\ge(1-\epsilon)\twonorm{\vct{w}^*}^2$. Thus, we have the following chain of inequalities
\begin{align}
\label{third3}
\frac{1}{n}\sum_{i=1}^n\abs{\langle \vct{x}_i,\vct{h}_{\perp}\rangle}^2\mathbb{1}_{\{\langle\vct{x}_i,\vct{w}\rangle\langle\vct{x}_i,\vct{w}^*\rangle\le 0\}}=&\frac{1}{n}\sum_{i=1}^n\abs{\langle\vct{x}_i,\vct{h}_{\perp}\rangle}^2\mathbb{1}_{\bigg\{\left(\frac{(\vct{w}^T\vct{w}^*)}{\twonorm{\vct{w}^*}^2}(\vct{x}_i^T\vct{w}^*)+\vct{x}_i^T\vct{h}_{\perp}\right)(\vct{x}_i^T\vct{w}^*)\le 0\bigg\}}\nonumber\\
=&\frac{1}{n}\sum_{i=1}^n\abs{\langle\vct{x}_i,\vct{h}_{\perp}\rangle}^2\mathbb{1}_{\bigg\{\frac{(\vct{w}^T\vct{w}^*)}{\twonorm{\vct{w}^*}^2}(\vct{x}_i^T\vct{w}^*)^2\le -(\vct{x}_i^T\vct{h}_{\perp})(\vct{x}_i^T\vct{w}^*)\bigg\}}\nonumber\\
\le&\frac{1}{n}\sum_{i=1}^n\abs{\langle\vct{x}_i,\vct{h}_{\perp}\rangle}^2\mathbb{1}_{\bigg\{\abs{\vct{x}_i^T\vct{w}^*}\le \twonorm{\vct{w}^*}^2\frac{\abs{\vct{x}_i^T\vct{h}_{\perp}}}{(\vct{w}^T\vct{w}^*)}\bigg\}}\nonumber\\
\le&\frac{1}{n}\sum_{i=1}^n\abs{\langle\vct{x}_i,\vct{h}_{\perp}\rangle}^2\mathbb{1}_{\bigg\{(1-\epsilon)\abs{\vct{x}_i^T\vct{w}^*}\le \abs{\vct{x}_i^T\vct{h}_{\perp}}\bigg\}}.
\end{align}
We now proceed by using the following result from \cite{soltanolkotabi2017structured}.
\begin{lemma}\cite{soltanolkotabi2017structured} Let $\mathcal{C}\in\R^d$ be a closed cone. Also let $\vct{w}^*$ be a fixed vector in $\R^d$. Furthermore, assume
\begin{align*}
n\ge c\cdot\omega^2(\mathcal{C}\cap\mathbb{S}^{d-1}),
\end{align*}
with $c$ a fixed numerical constant. Then 
\begin{align*}
\frac{1}{n}\sum_{i=1}^n\abs{\langle\vct{x}_i,\vct{h}_{\perp}\rangle}^2\mathbb{1}_{\bigg\{(1-\epsilon)\abs{\vct{x}_i^T\vct{w}^*}\le \abs{\vct{x}_i^T\vct{h}_{\perp}}\bigg\}}\le \left(\delta+\sqrt{\frac{21}{20}}\epsilon\right)^2\twonorm{\vct{h}}^2,
\end{align*}
holds with probability at least $1-2e^{-\gamma \delta^2 n}$ for all vectors $\vct{h}\in\mathcal{C}$ obeying
\begin{align*}
\twonorm{\vct{h}}\le \epsilon\twonorm{\vct{w}^*}.
\end{align*}
\end{lemma}
\begin{proof}
This lemma follows from the argument on pages 27-30 of \cite{soltanolkotabi2017structured}. 
\end{proof}
Combining the lemma with equations \eqref{third1}, \eqref{third2}, and \eqref{third3} we conclude that as long as $n\ge cn_0$, then
\begin{align}
\label{myinter3}
\frac{1}{n}\sum_{i=1}^n \left(1-\sgn{\langle \vct{x}_i,\vct{w}\rangle}\sgn{\langle \vct{x}_i,\vct{w}^*\rangle}\right)\sgn{\langle \vct{x}_i,\vct{w}^*\rangle}\langle \vct{x}_i,\vct{h}_{\perp}\rangle\langle\vct{x}_i,\vct{u}\rangle\le2\sqrt{1+\delta}\left(\delta+\sqrt{\frac{21}{20}}\epsilon\right)\twonorm{\vct{h}},
\end{align}
holds for all $\vct{u}\in\mathcal{C}\cap\mathbb{S}^{d-1}$ and $\vct{h}\in E(\epsilon)$ with probability at least $1-4e^{-\gamma n\delta^2}$. This completes the bound on the third term of \eqref{mymain}.

\subsubsection{Bounding the fourth term in \eqref{mymain}}
To bound the fourth term of \eqref{mymain} note that by the argument leading to \eqref{third1}
\begin{align}
\label{fourth1}
\frac{1}{n}\sum_{i=1}^n\bigg[&\frac{\sgn{\langle\vct{x}_i,\vct{w}\rangle}}{2}\left(1-\sgn{\langle\vct{x}_i,\vct{w}^*\rangle}\right)+\frac{\langle\vct{h},\vct{w}^*\rangle}{\twonorm{\vct{w}^*}^2}\bigg]\left(1-\sgn{\langle\vct{x}_i,\vct{w}\rangle}\sgn{\langle\vct{x}_i,\vct{w}^*\rangle}\right)\abs{\langle\vct{x}_i,\vct{w}^*\rangle}\langle\vct{x}_i,\vct{u}\rangle,\nonumber\\
&\le2\sqrt{1+\delta}\sqrt{\frac{1}{n}\sum_{i=1}^n\abs{\frac{\sgn{\langle\vct{x}_i,\vct{w}\rangle}}{2}\left(1-\sgn{\langle\vct{x}_i,\vct{w}^*\rangle}\right)+\frac{\langle\vct{h},\vct{w}^*\rangle}{\twonorm{\vct{w}^*}^2}}^2\abs{\langle\vct{x}_i,\vct{w}^*\rangle}^2\mathbb{1}_{\{\langle\vct{x}_i,\vct{w}\rangle\langle\vct{x}_i,\vct{w}^*\rangle\le 0\}}},\nonumber\\
&\le4\sqrt{1+\delta}\sqrt{\frac{1}{n}\sum_{i=1}^n\abs{\langle\vct{x}_i,\vct{w}^*\rangle}^2\mathbb{1}_{\{\langle\vct{x}_i,\vct{w}\rangle\langle\vct{x}_i,\vct{w}^*\rangle\le 0\}}},
\end{align}
holds for all $\vct{u}\in\mathcal{C}\cap\mathbb{S}^{d-1}$ with probability at least $1-2e^{-\gamma \delta^2n}$.
We again proceed by using the following result from \cite{soltanolkotabi2017structured}.
\begin{lemma}\cite{soltanolkotabi2017structured} Let $\mathcal{C}\in\R^d$ be a closed cone. Also let $\vct{w}^*$ be a fixed vector in $\R^d$. Furthermore, assume
\begin{align*}
n\ge c\cdot\omega^2(\mathcal{C}\cap\mathbb{S}^{d-1}),
\end{align*}
with $c$ a fixed numerical constant. Then 
\begin{align*}
\frac{1}{n}\sum_{i=1}^n\abs{\langle\vct{x}_i,\vct{w}^*\rangle}^2\mathbb{1}_{\{\langle\vct{x}_i,\vct{w}\rangle\langle\vct{x}_i,\vct{w}^*\rangle\le 0\}}\le \frac{1}{(1-\epsilon)^2}\left(\delta+\sqrt{\frac{21}{20}}\epsilon\right)^2\twonorm{\vct{h}}^2,
\end{align*}
holds with probability at least $1-2e^{-\gamma \delta^2 n}$ for all vectors $\vct{h}\in\mathcal{C}$ obeying
\begin{align*}
\twonorm{\vct{h}}\le \epsilon\twonorm{\vct{w}^*}.
\end{align*}
\end{lemma}
\begin{proof}
This lemma follows from the argument on pages 27-30 of \cite{soltanolkotabi2017structured}. 
\end{proof}
Combining the lemma with \eqref{fourth1} we conclude that as long as $n\ge cn_0$, then
\begin{align}
\label{myinter4}
\frac{1}{n}\sum_{i=1}^n \left(1-\sgn{\langle \vct{x}_i,\vct{w}\rangle}\sgn{\langle \vct{x}_i,\vct{w}^*\rangle}\right)\sgn{\langle \vct{x}_i,\vct{w}^*\rangle}\langle \vct{x}_i,\vct{h}_{\perp}\rangle\langle\vct{x}_i,\vct{u}\rangle\le\frac{4\sqrt{1+\delta}}{(1-\epsilon)^2}\left(\delta+\sqrt{\frac{21}{20}}\epsilon\right)\twonorm{\vct{h}},
\end{align}
holds for all $\vct{u}\in\mathcal{C}\cap\mathbb{S}^{d-1}$ and $\vct{h}\in E(\epsilon)$ with probability at least $1-4e^{-\gamma n\delta^2}$. This completes the bound on the fourth term of \eqref{mymain}.
\subsubsection{Putting the bounds together}
\label{all4}
In this Section we put together the bounds of the previous sections. Combining \eqref{myinter1}, \eqref{myinter2}, \eqref{myinter3}, and \eqref{myinter4} we conclude that
\begin{align*}
\langle \vct{u} ,\vct{w}-\vct{w}^*-\nabla \mathcal{L}(\vct{w})\rangle\le 2\left(\delta+\sqrt{1+\delta}\left(1+\frac{2}{(1-\epsilon)^2}\right)\left(\delta+\sqrt{\frac{21}{20}}\epsilon\right)\right)\twonorm{\vct{w}-\vct{w}^*},
\end{align*}
holds for all $\vct{u}\in\mathcal{C}\cap\mathbb{S}^{d-1}$ and $\vct{w}\in E(\epsilon)$ with probability at least $1-16e^{-\gamma \delta^2 n}-(n+10)e^{-\gamma n}$. Using this inequality with $\delta=10^{-4}$ and $\epsilon=7/200$ we conclude that
\begin{align*}
\langle \vct{u} ,\vct{w}-\vct{w}^*-\nabla \mathcal{L}(\vct{w})\rangle\le \frac{1}{4}\twonorm{\vct{w}-\vct{w}^*},
\end{align*}
holds for all $\vct{u}\in\mathcal{C}\cap\mathbb{S}^{d-1}$ and $\vct{w}\in E(\epsilon)$ with high probability.

\subsection*{Acknowledgements}
M.S. would like to thank Adam Klivans and Matus Telgarsky for discussions related to \cite{goel2016reliably} and the Isotron algorithm. This work was done in part while the author was visiting the Simons Institute for the Theory of Computing.

\bibliography{Bibfiles}
\bibliographystyle{plain}
\end{document}